\documentclass[letterpaper]{article} 
\usepackage{aaai25}  
\usepackage{times}  
\usepackage{helvet}  
\usepackage{courier}  
\usepackage[hyphens]{url}  
\usepackage{graphicx} 
\usepackage{natbib}  
\usepackage{caption} 
\usepackage{times}  
\usepackage{helvet}  
\usepackage{courier}  
\usepackage{algorithm}
\usepackage{algorithmic}
\usepackage{appendix}
\usepackage{amsmath}
\usepackage{threeparttable}
\usepackage{amsmath}
\usepackage{autobreak}
\usepackage{amsfonts}
\usepackage{mathrsfs}
\usepackage{booktabs}
\usepackage{graphicx} 
\usepackage{epstopdf}
\usepackage{subfigure}
\usepackage{multirow}
\usepackage{makecell}
\usepackage{caption}
\usepackage{textcomp,booktabs}
\usepackage[usenames,dvipsnames]{color}
\usepackage{colortbl}
\definecolor{mygray}{gray}{.9}
\definecolor{mypink}{rgb}{.99,.91,.95}
\definecolor{mycyan}{cmyk}{.3,0,0,0}
\usepackage{algorithm}
\usepackage{algorithmic}

\usepackage{newfloat}
\usepackage{listings}
\DeclareCaptionStyle{ruled}{labelfont=normalfont,labelsep=colon,strut=off} 
\floatstyle{ruled}
\newfloat{listing}{tb}{lst}{}
\floatname{listing}{Listing}
\title{Multimodal Hypothetical Summary for Retrieval-based \\ Multi-image Question Answering}
\author{
    Peize Li\textsuperscript{\rm 1, 2}\thanks{Work done during an internship at Institute of Information Engineering, Chinese Academy of Sciences.},
    Qingyi Si\textsuperscript{\rm 3},
    Peng Fu\textsuperscript{\rm 2, 4}\thanks{\rm Corresponding Authors. 
    },
    Zheng Lin\textsuperscript{\rm 2, 4},
    Yan Wang\textsuperscript{\rm 5, 1}\textsuperscript{\rm \dag}\\
    }
\affiliations{
    \textsuperscript{\rm 1}School of Artificial Intelligence, Jilin University, Changchun, China\\ 
    \textsuperscript{\rm 2}Institute of Information Engineering, Chinese Academy of Sciences, Beijing, China\\
    \textsuperscript{\rm 3}Huawei Technologies Co., Ltd., Beijing, China\\    
    \textsuperscript{\rm 4}School of Cyber Security, University of Chinese Academy of Sciences, Beijing, China\\ 
    \textsuperscript{\rm 5}Key Laboratory of Symbol Computation and Knowledge Engineering of Ministry of Education, College of Computer Science and Technology, Jilin University, Changchun, China\\    
    lipz21@mails.jlu.edu.cn, siqingyi@huawei.com,
    \{fupeng, linzheng\}@iie.ac.cn,
    wy6868@jlu.edu.cn \\
    
}

\usepackage{bibentry}

\begin{document}

\maketitle

\begin{abstract}
Retrieval-based multi-image question answering (QA) task involves retrieving multiple question-related images 
and synthesizing these images to generate an answer. 
Conventional ``retrieve-then-answer" pipelines often suffer from cascading errors
because the training objective of QA fails to optimize the retrieval stage. 
To address this issue, 
we propose a novel method to effectively introduce and reference retrieved information into the QA. 
Given the image set to be retrieved,
we employ
a multimodal large language model (visual perspective) and a large language  model (textual perspective) 
to obtain
\textit{multimodal hypothetical summary} in question-form and description-form. 
By combining visual and textual perspectives, 
MHyS captures image content more specifically 
and replaces \textit{real} images in retrieval,
which eliminates the modality gap by transforming into text-to-text retrieval
and helps improve retrieval.
To more advantageously introduce retrieval with QA, 
we employ contrastive learning to align queries (questions) with MHyS.
Moreover, 
we propose a coarse-to-fine strategy for calculating both sentence-level and word-level similarity scores, 
to further enhance retrieval and filter out irrelevant details. 
Our approach achieves a 3.7$\%$ absolute improvement over state-of-the-art methods 
on RETVQA and a 14.5$\%$ improvement over CLIP. 
Comprehensive experiments and detailed ablation studies demonstrate the superiority of our method.
\end{abstract}

%

\section{Introduction}
\begin{figure}[t]
    \centering
    \begin{center}
    \includegraphics[width=3.2in]{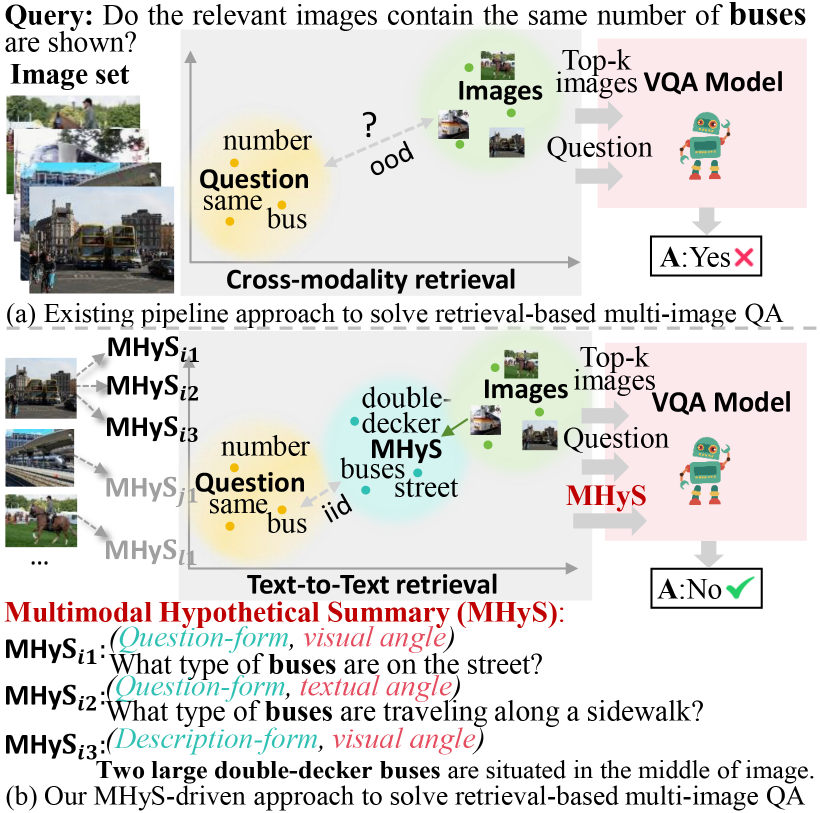}
    \end{center}
    \caption{An illustration of our motivation. 
    Compared to the ``retrieve-then-answer" pipeline, our approach leverages multimodal hypothetical summary (MHyS) to transform cross-modality retrieval into text-to-text retrieval, 
    effectively introducing and referencing retrieval into QA.
    }
    \label{fig:motivation}
\end{figure}
Visual question answering (VQA) is an engaging multimodal task that involves analyzing both images and natural language. 
Unlike conventional VQA tasks, retrieval-based multi-image QA \cite{penamakuri2023answer} 
is a newly proposed and more challenging VQA task.
It requires retrieving and integrating multiple question-related images 
to generate an answer.
Existing vision-language models often rely on extensive pre-training \cite{li2022mplug,cho2021unifying,si2023combo} or sophisticated attention mechanisms \cite{kim2018bilinear,anderson2018bottom,zhu2021mucko} to tackle multimodal tasks. 
However, they face difficulties with retrieval-based multi-image QA due to the limited exposure to multi-image  question-answering pairs during pre-training and the added complexity of the retrieval process.

To advance research in this area, recent work has proposed several solutions. 
For example, REALM \cite{guu2020retrieval}, RAG \cite{lewis2020retrieval} and RETRO \cite{borgeaud2022improving} utilize extensive world knowledge from language models to improve QA performance, though they are limited to textual data. 
To broaden the scope of knowledge, MuRAG \cite{chen2022murag} extends these approaches to multimodal corpora. 
Besides,  Solar \cite{yu2023unified} transforms images and tables into a unified text format, 
utilizing the advantages of language models more effectively. 
Despite enhancing knowledge and content generation, 
these methods are still part of the ``retrieve-then-answer" pipeline, 
and face the challenge of misaligned training objectives between retrieval and QA.
To address this issue,
our method employs multimodal hypothetical summary to replace images during retrieval, 
optimizing the network end-to-end using contrast enhancement loss and VQA loss.
We illustrate this idea with the example shown in Figure \ref{fig:motivation}.

This example reflects the two purposes of our designed MHyS, 
encompassing both question-form and description-form.
First, for the query, ``\textit{Do the relevant images contain the same number \textbf{buses} are shown?}",
MHyS captures important details effectively, such as ``\textit{\textbf{buses}}".  
Besides, the question-form MHyS aligns with query in terms of semantic overlap (same words) and structural similarity (question format).
Second, 
the description-form MHyS provides description-relevant visual information,
such as ``\textit{\textbf{Two large double-decker buses}}", 
which not only aids in retrieval but also helps the model identify question-attended evidence.

The ``retrieve-then-answer" pipelines \cite{penamakuri2023answer,chen2022murag,yu2023unified}
first utilize cross-modality retrieval method (e.g. BLIP \cite{li2022blip}, ALBEF \cite{li2021align} and mPLUG \cite{li2022mplug}) to rank images based on question.
Then, these question-related images are combined with question into VQA model 
(e.g. BAN \cite{kim2018bilinear}, LXMERT \cite{tan2019lxmert}, VL-BART \cite{cho2021unifying}).
Instead of decomposing retrieval-based multi-image QA task,  
our MHyS-driven approach uses multimodal hypothetical summary to connect retrieval and QA.
Concretely, we replace real images with MHyS to calculate the similarity score with query,
effectively transforming text-to-text retrieval. 
To retain more information and contact retrieval with QA,
we combine the selected real images (based on similarity scores) with their MHyS to generate answers.

We expect the multimodal hypothetical summary (MHyS) to proficiently capture key object words (overlapping with query words) and provide more specific information about real images, enhancing both retrieval and QA. 
To achieve this, we design question-form and description-form MHyS. 
To effectively utilize vision-language alignment knowledge and filter out irrelevant details, 
we employ the frozen CLIP \cite{radford2021learning} 
(pre-trained on 400M image-text pairs using contrastive learning) 
to calculate
sentence-level similarity 
and a multimodal encoder \cite{cho2021unifying} 
(rich in multimodal knowledge) 
to compute word-level similarity. 
We adopt a coarse-to-fine strategy for accurate similarity calculation. 
To seamlessly introduce and effectively utilize 
retrieval and QA, 
we use contrastive enhancement loss to align query with MHyS. 
This approach also enables the model to capture more generalized representation.

The main contributions are summarized as follows:
(1) We propose an innovative MHyS-driven paradigm that effectively transforms into 
text-to-text retrieval 
and seamlessly connects retrieval with QA.
(2) We design the question-form MHyS to closely align with query through semantic overlap and similar structure, 
while the description-form MHyS not only facilitates retrieval but also helps in capturing question-attended evidence.
(3) The proposed multi-granularity retrieval aims to precisely filter out irrelevant details in coarse-to-fine strategy.
(4) Our approach achieves a 3.7$\%$ absolute accuracy improvement over the state-of-the-art on the RETVQA dataset and a 14.5$\%$ enhancement over CLIP, demonstrating its effectiveness.

\section{Related Work} \label{Related Work}
\begin{figure*}[ht]
    \centering
    \begin{center}
    \includegraphics[width=7.0in]{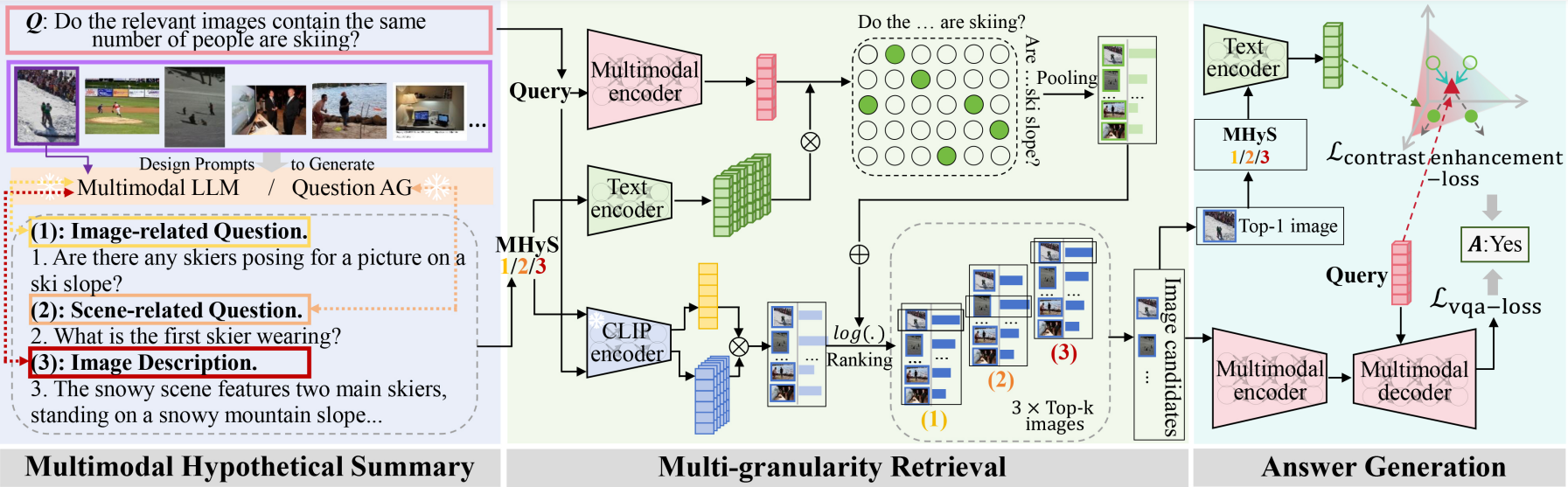}
    \end{center}
    \caption{The overview of our approach. 
    \textbf{Multimodal Hypothetical Summary} (MHyS) employs
    multimodal 
    large language model (visual perspective) and language large model (textual perspective) to 
    obtain  
    both question-form and description-form \textit{hypothetical} summary, 
    which replaces \textit{real} images during retrieval 
    and eliminates the modality gap by transforming into text-to-text retrieval. 
    \textbf{Multi-granularity Retrieval} calculates sentence-level and word-level similarities to rank images. 
    To capture more information, 
    the selected \textit{real} images (based on similarity scores) are combined with their MHyS to generate the answers.
    }
    \label{fig:model}
\end{figure*}

\subsubsection{Visual Question Answer and Retrieval-based Multi-image QA.}
In recent years, various interesting studies have been proposed for visual question answering (VQA). 
Conventional VQA tasks,
such as  knowledge-based VQA \cite{marino2019ok,schwenk2022okvqa}, visual reasoning VQA \cite{hudson2019gqa, zellers2019recognition} and
language bias VQA \cite{agrawal2018don,si2022language},
have been thoroughly studied. 
Retrieval-based QA tasks currently include the WebQA \cite{chang2022webqa} and RETVQA \cite{penamakuri2023answer}.
WebQA relies on retrieving textual knowledge and has been addressed by various LLM-based methods \cite{chen2022murag,yu2023unified,gu-etal-2024-light}. 
However, RETVQA (retrieval-based multi-image question answering) is newly proposed and more challenging in comparison.
It requires retrieving multiple question-related images from a collection of relevant as well as irrelevant images.
The challenge lies in accurately retrieving question-relevant images and providing the correct answer.
Existing solution MI-BART \cite{penamakuri2023answer} for this task uses a two-stage ``retrieve-then-read" pipeline.
Unlike separating the task into two sub-tasks, our method uses MHyS to introduce multi-image retrieval into QA,
which gets rid of the cascading error in two-stage pipelines.

\subsubsection{Image-to-Text Retrieval.}
Previous methods typically improve retrieval performance in two ways: 
one is to pretrain on larger datasets and 
the other is to optimize the retrieval components.
For example, widely used multimodal 
models like BLIP \cite{li2022blip} and mPLUG \cite{li2022mplug} are trained on 14M caption-based datasets including  MSCOCO \cite{lin2014microsoft} and Flickr30K \cite{young2014image}, using Image-Text Contrastive (ITC) and Image-Text Matching (ITM) losses. 
These models, which operate at the word level, excel in aligning entity semantics. 
Benefit from contrastive learning, 
CLIP \cite{radford2021learning} contains vast vision-language alignment knowledge (pre-trained on 400 million image-text pairs) to select important visual information.
Therefore, CLIP excels at sentence-level encoding.
Our approach combines the strengths of both methods by integrating sentence-level and word-level similarity for more accurate image-text matching.

Typical retrieval methods mainly focus on query expansion or re-ranking scheme. 
For example, DQU-CIR \cite{wen2024simple} creates a unified multimodal query by merging visual and textual queries, while LeaPRR \cite{qu2023learnable} follows the latter way
and uses graph reasoning to explore higher-order intra- and inter-modal relationships. 
These approaches require time-consuming pre- and post-processing. 
Without requiring any extra annotations,
our approach utilizes MHyS (question-form and description-form) to transform 
into text-to-text retrieval,
effectively introducing and referencing retrieved information into QA.

Recently, advanced RAG \cite{eibich2024aragog} method uses large language models (LLMs) to rewrite queries.
The multimodal RAG RA-CM3 \cite{yasunaga2023retrieval} can retrieve and generate both text and images. 
In addition to deriving the corpus from external documents, the corpus can also be artificially constructed. 
Hypothetical Document Embedding \cite{gao2023precise} enhances document retrieval by using LLM to generate a hypothetical answer to a query.
However, it may produce hallucinated queries that introduce noise and is limited to text retrieval.
On the contrary, our method uses multimodal hypothetical summary (MHyS) to replace images being retrieved rather than queries,
helping align queries with target images and capture question-attended valuable evidence for QA.

\section{Methodology}
Given the  retrieval-based multi-image dataset 
$\mathcal{D}=\{(\boldsymbol{Q}_i, \mathcal{I}_{i}, \boldsymbol{A}_i)\}_{i=1}^{N}$ 
with $N$ samples,
where $\boldsymbol{Q}_i$,  $\mathcal{I}_{i}$ and $\boldsymbol{A}_i$ denote the question, image set and ground-truth answer of $i$-th sample respectively. 
The image set $\mathcal{I}_{i}=\{\boldsymbol{I}_{i1},\boldsymbol{I}_{i2},...,\boldsymbol{I}_{ij}\}_{j=1}^{N_{e}}$  includes both relevant and irrelevant images to the question, 
the task is to select multiple question-related images from the image set $\mathcal{I}_{i}$ and 
use these selected images, along with the question, to generate the answer $\boldsymbol{A}_i$.
Figure \ref{fig:model} illustrates an overview of our method, encompassing multimodal hypothetical summary, multi-granularity retrieval and answer generation.

\subsection{Multimodal Hypothetical Summary}
Multimodal hypothetical summary (MHyS) aims to obtain question-form and description-form 
multimodal hypothetical summary, 
which aids in aligning with query and provides question-attended valuable evidence for QA.
By replacing real images in retrieval,
MHyS mitigates inter-modal out-of-domain issue by transforming into text-to-text retrieval.

Concretely, given the $i$-th sample ($\boldsymbol{Q}_i$, $\mathcal{I}_{i}$, $\boldsymbol{A}_i$), 
we obtain MHyS for each image $\boldsymbol{I}_{ij} \in \mathcal{I}_{i}$ using the following approach:
(1) Image-related Question: 
We utilize the \textit{Multimodal Large Language Model} mPLUG-Owl2 \cite{ye2024mplug} with the prompt ``Generate a question based on the image:" to create MHyS $\boldsymbol{Q}_{ij}^{m}$ in the form of question. 
This approach directly generates MHyS based on images (visual perspective), 
which focuses more on the visual content. 
mPLUG-Owl2 \cite{ye2024mplug} is a multimodal LLM with strong multimodal generation capabilities and proficiency in visual reasoning.
The multimodal LLM, which is further trained to follow instructions, can zero-shot generalize to diverse prompts.
(2) Scene-related Question: 
To make MHyS diverse,
we propose question-form MHyS from textual perspective. 
First, we use mPLUG-owl2 \cite{ye2024mplug} to generate scene information with the prompt ``Generate the scene information based on the image:". 
The scene information is then fed into \textit{Question and Answer Generation model} QAG \cite{ushio2023empirical} model to generate scene-related questions $\boldsymbol{Q}_{ij}^{d}$. 
This MHyS enriches the visual perspective with textual perspective.
(3) Image Description: 
Question-form MHyS may not cover specific image information.
To address this, 
we use mPLUG-Owl2 to generate detailed image descriptions $\boldsymbol{D}_{ij}$ based on images with the prompt ``Generate a detailed description based on the image:".
This process (1)-(3) generates multifaceted MHyS that integrate both visual and textual perspectives,
enhancing both retrieval and QA.
Following TwO \cite{si2023combo}, 
we use 600-dimensional word embeddings \cite{pennington2014glove} in conjunction with GRU to obtain the base embedding representations ($\boldsymbol{Q}_{ij}^{m}$, $\boldsymbol{Q}_{ij}^{d}$, $\boldsymbol{D}_{ij}$).

The question-form MHyS matches the query in structure (both being in question form) and semantics (sharing the same keywords). 
Meanwhile, the description-form MHyS includes not only keywords but also more specific visual details that support QA task. 
Together, these components greatly enhance the overall performance of the method.

\subsection{Multi-granularity Retrieval}
To more precisely filter out irrelevant details and improve retrieval accuracy, 
we employ a coarse-to-fine strategy to compute similarity scores at both the sentence and word levels.

To achieve global alignment, 
we leverage CLIP \cite{radford2021learning} for sentence-level encoding and similarity calculation, 
which has been pre-trained on 400 million image-text pairs using contrastive learning and possesses a strong implicit cross-modal alignment capability.
Specifically, we use CLIP text encoder to process the query (question $\boldsymbol{Q}_i$)
and MHyS components 
(image-related question $\boldsymbol{Q}_{ij}^{m}$, scene-related question $\boldsymbol{Q}_{ij}^{d}$ and image description $\boldsymbol{D}_{ij}$), 
as illustrated below:
\begin{equation} \label{eq1}
\boldsymbol{T}_i={\rm CLIP_{text}}(\boldsymbol{Q}_i)
\small
\end{equation}
\begin{equation} \label{eq2}
\boldsymbol{\hat {Q}}_{ij}^{m}, \boldsymbol{\hat {Q}}_{ij}^{d}, \boldsymbol{\hat {D}}_{ij}={\rm CLIP_{text}}(\boldsymbol{Q}_{ij}^{m}, \boldsymbol{Q}_{ij}^{d}, \boldsymbol{D}_{ij})
\small 
\end{equation}

To facilitate word-level semantic alignment between the query and MHyS 
while filtering out irrelevant details, 
we employ a multimodal encoder for word-level encoding and similarity calculation. 
Concretely,  we adopt the multimodal encoder VL-BART \cite{cho2021unifying} to encode query tokens:
\begin{equation} \label{eq3}
\boldsymbol{\overline T}_i={\rm Enc_{multi}}(\boldsymbol{Q}_i)
\small
\end{equation}
To encode MHyS at the word level, 
we use a new textual encoder comprising six transformer layers:
\begin{equation} \label{eq4}
\boldsymbol{\overline {Q}}_{ij}^{m}, \boldsymbol{\overline {Q}}_{ij}^{d}, \boldsymbol{\overline {D}}_{ij}={\rm Enc_{text}}(\boldsymbol{Q}_{ij}^{m}, \boldsymbol{Q}_{ij}^{d}, \boldsymbol{D}_{ij})
\small
\end{equation}
Each transformer layer consists of a self-attention layer followed by a fully connected linear layer with residual connections.
We implement this way because the multimodal encoder 
maintains a cohesive encoding space for questions and images, enhancing the effectiveness of visual question answering.

Next, we compute similarity scores between the query and MHyS (using image-related questions $\boldsymbol{Q}_{ij}^{m}$ as an example) at both the sentence-level and word-level:
\begin{equation} \label{eq5}
s(\boldsymbol{T}_i,\boldsymbol{\hat {Q}}_{ij}^{m})=f_q(\boldsymbol{T}_i)^{T}f_v(\boldsymbol{\hat {Q}}_{ij}^{m})  \in \mathbb{R}^{1\times 1}
\small
\end{equation}
\begin{equation} \label{eq6}
s(\boldsymbol{\overline T}_i,\boldsymbol{\overline {Q}}_{ij}^{m})=g_q(\boldsymbol{\overline T}_i)^{T}g_v(\boldsymbol{\overline {Q}}_{ij}^{m}) \in \mathbb{R}^{N_l\times N_v}
\small
\end{equation}
where $s(.)$ represents the cosine similarity function.
$f_q$, $f_v$, $g_q$ and $g_v$ are the multi-layer perceptron (MLPs) used to encode features. 
$N_l$ denotes the number of query words, and $N_v$ represents the number of textual words in the MHyS.

Guided by the relevance affinity matrix, we identify the most relevant MHyS content for the query. 
The multimodal and textual encoders capture the implicit correlations among all MHyS worlds, 
and the selected query-centric MHyS world incorporates its contextual information, 
providing crucial clues for retrieval.
Specifically, we use max-pooling to assess the relevance of each MHyS world to the query as follows:
\begin{equation} \label{eq7}
\widetilde s(\boldsymbol{\overline T}_i,\boldsymbol{\overline {Q}}_{ij}^{m})={\mathop {\max}\limits_{N_v}}({\mathop {\max}\limits_{N_l}}(s(\boldsymbol{\overline T}_i,\boldsymbol{\overline {Q}}_{ij}^{m})))) \in \mathbb{R}^{1\times 1}
\small
\end{equation}

To enhance retrieval accuracy, we integrate both sentence-level and word-level similarities:
\begin{equation} \label{eq8}
h(\boldsymbol{Q}_i,\boldsymbol{Q}_{ij}^{m})=\widetilde s(\boldsymbol{\overline T}_i,\boldsymbol{\overline {Q}}_{ij}^{m})+log(s(\boldsymbol{T}_i,\boldsymbol{\hat {Q}}_{ij}^{m}))
\small
\end{equation}
To address discrepancies in data magnitudes, we apply a $log(.)$ transformation to the word-level similarity before combining it with the sentence-level similarity. 
Besides, we explore various similarity fusion functions, including direct addition and adaptive methods, which are discussed in detail in the ablation experiments.

We rank the candidate images based on the similarity score of MHyS (image-related question $\boldsymbol{Q}_{ij}^{m}$), as follows:
\begin{equation} \label{eq9}
\begin{split} 
\mathcal{C}_{qm} =  topK(argsort(h(\boldsymbol{Q}_i,\boldsymbol{Q}_{ij}^{m})))  \\
\end{split}
\small
\end{equation}

Similar to equations 5-8,
we compute similarity scores and identify candidate images for MHyS (e.g., scene-related questions $\boldsymbol{Q}_{ij}^{d}$) and  MHyS (e.g., image descriptions $\boldsymbol{D}_{ij}$):
\begin{equation} \label{eq10}
h(\boldsymbol{Q}_i,\boldsymbol{Q}_{ij}^{d})=\widetilde s(\boldsymbol{\overline T}_i,\boldsymbol{\overline {Q}}_{ij}^{d})+log(s(\boldsymbol{T}_i,\boldsymbol{\hat {Q}}_{ij}^{d}))
\small
\end{equation}
\begin{equation} \label{eq11}
\begin{split} 
\mathcal{C}_{qd} =  topK(argsort(h(\boldsymbol{Q}_i,\boldsymbol{Q}_{ij}^{d})))  \\
\end{split}
\small
\end{equation}
\begin{equation} \label{eq12}
h(\boldsymbol{Q}_i,\boldsymbol{D}_{ij})=\widetilde s(\boldsymbol{\overline T}_i,\boldsymbol{\overline {D}}_{ij})+log(s(\boldsymbol{T}_i,\boldsymbol{\hat {D}}_{ij}))
\small
\end{equation}
\begin{equation} \label{eq13}
\begin{split} 
\mathcal{C}_{d} =  topK(argsort(h(\boldsymbol{Q}_i,\boldsymbol{D}_{ij})))  \\
\end{split}
\small
\end{equation}

By combining candidate images from these three types of MHyS, 
we generate the final set of candidate images 
$\mathcal{C}_{i}$ for each multi-image QA pair ($\boldsymbol{Q}_i$, $\mathcal{I}_{i}$, $\boldsymbol{A}_i$), 
as shown:
\begin{equation} \label{eq14}
\mathcal{C}_{i} = \mathcal{C}_{qm} \cup \mathcal{C}_{qd} \cup \mathcal{C}_{d}
\end{equation}

\subsection{Answer Generation}
To more effectively introduce and reference retrieved information into QA, 
we employ contrastive learning to align query with MHyS
and VQA loss to align question with 
question-focused visual content.

To calculate contrast enhancement loss, for the $i$-th sample, 
we use the query feature $\boldsymbol{\overline T}_i$
processed by multimodal encoder and the top-1 retrieved MHyS feature $\boldsymbol{\overline {Q}}_{ij}^{m}$ processed by text encoder.
We define positive samples $(\boldsymbol{\overline T}_i, \boldsymbol{\overline Q}_i^{m+})$ and negative sample pairs 
$(\boldsymbol{\overline T}_b, \boldsymbol{\overline Q}_b^{m})_{b=1}^{B}$
within the same batch. 
($b \neq i$). 
$B$ denotes the number of negative samples in a batch. 
Following the MMBS \cite{si2022towards} study, 
we use the cosine similarity function to compute the contrastive enhancement loss (in short $\mathcal{L}_{CE}$), 
formulated as follows:
\begin{equation} \label{eq15}
\mathcal{L}_{CE}= -\log\frac{e^{cos(\boldsymbol{\overline T}_i, \boldsymbol{\overline Q}_i^{m+})}}{e^{cos(\boldsymbol{\overline T}_i, \boldsymbol{\overline Q}_i^{m+})}+\sum\limits_{b=1}^B {e^{cos(\boldsymbol{\overline T}_b, \boldsymbol{\overline Q}_b^{m})}}}
 \small
\end{equation}

To calculate VQA loss, 
we utilize the multimodal encoder VL-BART to encode the retrieved images 
$\mathcal{C}_{i} = \{\boldsymbol{I}_{i1}, \boldsymbol{I}_{i2},...,\boldsymbol{I}_{ij}\}$,
as illustrated below:
\begin{equation} \label{eq16}
\boldsymbol{\overline I}_{i1}, \boldsymbol{\overline I}_{i2}, ..., \boldsymbol{\overline I}_{ij}={\rm Enc_{multi}}(\boldsymbol{I}_{i1}, \boldsymbol{I}_{i2},...,\boldsymbol{I}_{ij})
\small
\end{equation}
The encoded images 
$(\boldsymbol{\overline I}_{i1}, \boldsymbol{\overline I}_{i2}, ..., \boldsymbol{\overline I}_{ij})$ 
and question ${\overline T}_{i}$ 
are fed into 
VL-BART decoder 
to generate the final answer according to 
the prediction probability $P(.)$ over the vocabulary
space $|W|$ for each answer token:
\begin{equation} \label{eq17}
P(a_i ^1), ...,P(a_i ^l) = Softmax({\rm Dec_{multi}}({\overline T}_{i}, \boldsymbol{\overline I}_{i1}, \boldsymbol{\overline I}_{i2}, ..., \boldsymbol{\overline I}_{ij}))
\small
\end{equation}
We minimize the auto-regressive cross-entropy loss
for QA:
\begin{equation} \label{eq18}
L_{VQA} =\frac{-1}{N \cdot L \cdot |W|} \sum\limits_{i=1}^N \sum\limits_{l=1}^L \sum\limits_{w=1}^{|W|} A_i ^{l,w} log(P(a_i ^{l,w}))
\small
\end{equation}
where $l$ represents the answer length and $N$ 
denotes the number of samples.

\begin{table*}[ht]
  \centering
  \fontsize{300pt}{300pt}\selectfont
  \resizebox{1.4\columnwidth}{!}{
    \begin{tabular}{l|c|ccccc}
    \toprule
    Models & All & Attribute  & Color & Count & Relation & Shape \\
    \midrule
    \rowcolor{mygray}
    \textit{Introduce  retrieval into QA} &    &       &       &       &       &  \\    
    CLIP \cite{radford2021learning}-BAN \cite{kim2018bilinear}  & 14.4 & 0.0 & 17.6 & 0.3 & 13.6 & 37.0 \\
    CLIP \cite{radford2021learning}-VisualBERT \cite{li2019visualbert}  & 19.2  &  0.0 &  26.6  & 0.4 & 23.2    & 44.4 \\   
    CLIP \cite{radford2021learning}-LXMERT \cite{tan2019lxmert} &  21.9 & 0.0 & 32.5 & 0.4 & 30.8 & 47.2 \\ 
    CLIP \cite{radford2021learning}-VLBART \cite{cho2021unifying}  & 65.8 & 78.9 & 66.4 & 55.3 & 15.6 & 83.9 \\       
    \hline
    \rowcolor{mygray}
    \textit{Retrieve then answer} &      &       &       &       &       &  \\
    mPLUG \cite{li2022mplug}-BAN \cite{kim2018bilinear}&  15.2 & 0.0 & 19.4 & 0.3 & 14.8 & 38.5 \\    
    mPLUG \cite{li2022mplug}-VisualBERT \cite{li2019visualbert}&  19.1&  0.0 & 28.1 &0.3& 21.3 & 44.0 \\ 
    mPLUG \cite{li2022mplug}-LXMERT \cite{tan2019lxmert} &   19.7 &0.0     & 32.4 & 0.3 & 20.5 &  42.1\\   
    MI-VLBART \cite{cho2021unifying} (Question only) & 62.4  & 74.9 & 58.0 & 51.6 & 12.4 & 86.9 \\
    MI-VLP \cite{zhou2020unified} &  65.1  & 76.8 & 62.0 & 50.8 & 36.8 & 84.0 \\    
    MI-(LSTM+VGG) \cite{antol2015vqa} (Aggregate VQA)&   66.6  & 75.4 & 60.1 & 54.6 & 32.2 & 91.3 \\
    MI-BART \cite{penamakuri2023answer} (Image stitch variant) & 72.1  & \textbf{81.6} & 71.8 & 62.7 & 52.0 & 96.2 \\
    mPLUG \cite{li2022mplug}-VLBART \cite{cho2021unifying} & 75.3 & 76.6 & 69.6 & 80.2 & 33.1 & 92.3 \\
    MI-BART \cite{penamakuri2023answer}  & 76.5  & 78.5 & 72.1 & 66.0 & \textbf{69.5} & 92.4 \\ 
    \midrule
    \rowcolor{mygray}
    \textbf{Ours} \textit{Introduce retrieval into QA} & \textbf{80.3} & 78.5 & \textbf{79.5} & \textbf{80.5} & 40.4 & \textbf{97.9} \\
    \bottomrule
    \end{tabular}%
    }
    \caption{Comparison with state-of-the-art approaches on the retrieval-based multi-image QA dataset using the retrieved images. 
    ``\textit{retrieve then answer}" pipeline involves two separating stages: first, cross-modality retrieval and then VQA-based answer prediction.
    In contrast, ``\textit{Introduce retrieval into QA}" approach adopts our proposed MHyS-driven paradigm, 
    which replaces images with multimodal hypothetical summary (MHyS) for retrieval, and selects the images (based on similarity scores) along with their MHyS to generate answers.
    The hyphen ``-" before and after denote the retrieval method and VQA model respectively.
    }
  \label{tab:addlabel1}%
\end{table*}%

\section{Experiments}

\subsection{Dataset and Experimental Settings}
\subsubsection{Dataset.} We evaluate our approach on the retrieval-based multi-image QA dataset \cite{penamakuri2023answer}, which contains 334K samples for training, 41K for validation and 41K for testing. 
The questions cover various types, including color, shape, counting, object attributes and relations. 
To further test the generalization of our method, we retrieve the top 1-10 images to evaluate performance.

\subsubsection{Experimental settings.}
We conduct our experiments using an NVIDIA 3090 24GB GPU. 
We train the network for 20 epochs with the batch size of 100 and an initial learning rate of 1e-4, using AdamW optimizer.
The text encoder has a dimension of 768.
Accuracy verifies whether the correct answer is among the generated answers, consistent with the original dataset proposed.
We adopt the CLIP model with RN50$\times$64 visual encoder backbone.
To ensure reliable results, we provide the average performance from three trials.

\subsection{Comparisons with State-of-the-Arts}
Table \ref{tab:addlabel1}  shows the comparison of
our method with state-of-the-art (SoTA) models,
categorized into ``retrieve then answer" and ``introduce retrieval into QA" 
based on whether the retrieval process is introduced into the QA process.
Several observations can be derived: 
(1) Two-stage ``retrieve then answer" approaches generally perform better among the methods compared. 
However, our approach, which introduces retrieval into the QA process, surpasses the state-of-the-art two-stage method MI-BART by +3.7$\%$. 
MI-BART uses a cross-modality relevance retriever pre-trained on COCO and fine-tuned on RETVQA, coupled with a transformer-based encoder-decoder framework in the QA phase. 
By leveraging the Multimodal Hypothetical Summary (MHyS),  
our approach more effectively introduces and references the retrieved information into the QA process, resulting in improved performance. 
(2) Compared to ``Introduce  retrieval into QA" methods, our approach exceeds the SoTA CLIP-BART by +14.5$\%$. 
By using MHyS to  replace  visual images in retrieval, 
we shift to text-to-text retrieval, 
effectively aligning with queries and capturing question-attended valuable evidence for QA.

\subsection{Ablation Study}
\subsubsection{Ablation of Multimodal Hypothetical Summary.}

\begin{table}[t]
  \centering
  \fontsize{80pt}{80pt}\selectfont
  \resizebox{1.0\columnwidth}{!}{  
    \begin{tabular}{l||c|ccccc}
    \toprule
    Models & All & Att  & Col & Cou & Rel & Sha \\
    \midrule
    MHyS-Caption & 69.82 & 78.13 & 68.60 & 60.83 & 25.87 & 90.47 \\
    MHyS-Description & 70.87 & \textbf{79.18} & 70.87 & 61.54 & 25.11 & 90.55 \\
    MHyS-(Image-Ques) & 71.56 &79.00  & 69.45 & 61.86 & 26.64 & 91.51 \\  
    MHyS-(Scene-Ques) & 73.41 & 77.69 &72.52  & 67.17 & 32.17 & 93.15 \\  
    MHyS-Question & 78.45 & 77.85 & 77.82 & 76.01 & 39.19 & 96.47 \\
    MHyS-(Desc+Ques) w/ RA-loss & 79.81 & 78.41 & 78.87 & 78.57 & 40.25 & 97.71 \\     
    MHyS-(Desc+Ques) (\textbf{Ours}) & \textbf{80.30} & 78.51 & \textbf{79.45} & \textbf{80.52} & \textbf{40.44} & \textbf{97.85} \\   
    \bottomrule
    \end{tabular}%
    }
    \caption{Ablation study of multimodal hypothetical summary.}
  \label{tab:multimodal hypothetical summary}%
\end{table}%

We evaluate various aspects and specific forms of MHyS in Table \ref{tab:multimodal hypothetical summary}
and derive several key findings:
(1) Question-form MHyS proves to be the most effective, 
outperforming Description-form MHyS by +7.58$\%$. 
It  aligns closely with the question-form query and includes key visual object words, such as ``bus" in Figure \ref{fig:motivation}.
(2) Developed from textual perspective by the QAG model, MHyS-(Scene-Ques) surpasses MHyS-(Image-Ques) by +1.85$\%$. 
The latter is generated from visual perspective by the MLLM.
Combining these two approaches results in a +8.25$\%$ performance increase, emphasizing the benefit of obtaining MHyS from both visual and textual perspectives.
(3)  MHyS-Description performs slightly better than MHyS-Caption because it covers more scene details. 
Combining MHyS-Question and MHyS-Description results in a +9.99$\%$  overall improvement in our approach. 
(4) Moreover, 
contrastive enhancement loss provides a +0.49$\%$ increase in performance by further aligning the query with MHyS.

\subsubsection{Ablation of Retrieval Similarity.}
\begin{table}[t]
  \centering
  \fontsize{30pt}{30pt}\selectfont
  \resizebox{1.0\columnwidth}{!}{
    \begin{tabular}{l|c|ccccc}
    \toprule
    Models & All & Att & Col & Cou & Rel & Sha \\
    \midrule
    CLIP (\textit{sentence-level}) &   63.69    &  79.15     & 62.11      &  52.21     &  12.76     & 84.17 \\
    $\boldsymbol \log$ (CLIP)+$\boldsymbol \log$(MHyS)  & 64.23 & 78.63& 64.16 & 53.70 & 14.70 & 81.75 \\
    CLIP+$\boldsymbol \log$  & 73.66 & 79.41 & 71.42 & 66.36 & 33.79 & 93.32 \\
    CLIP+MHyS & 73.86 & \textbf{79.83} & 71.99 & 67.13 & 34.62 & 93.15 \\
    Adap-Func [CLIP;MHyS] & 80.03 & 78.78 & 79.22 & 79.61 & 39.29 & 97.55 \\
    MHyS (\textit{world-level}) & 80.09 & 78.45 & 79.45 & 79.52 & 39.44 & 97.85 \\
    CLIP+$\boldsymbol \log$(MHyS)  (\textbf{Ours}) & \textbf{80.30} & 78.51 & \textbf{79.45} & \textbf{80.52} & \textbf{40.44} & \textbf{97.85} \\
    \bottomrule
    \end{tabular}%
    }
  \caption{Ablation study of similarity calculation methods.}
  \label{tab:similarity}%
\end{table}%

We assess the
the effect of various similarity computation strategies in Table \ref{tab:similarity}.
An optimal retrieval similarity method should be both simple and effective.
We observe the following:
(1) \textit{Fine-grained similarity is far more effective than coarse-grained similarity}.
For example, ``CLIP (sentence-level)" uses global similarity based on entire sentence and image, 
while ``MHyS (word-level)" measures similarity at the object and word level. 
MHyS (word-level)  significantly outperforms CLIP (sentence-level) by +16.4$\%$.
(2) \textit{Properly combining coarse-grained and fine-grained methods is crucial}.
We experiment with direct addition, $\boldsymbol \log(.)$ transformation, and training a fusion adaptive function. 
The best result is achieved by combining addition with
$\boldsymbol \log(.)$ transformation,
which balances magnitude differences,  
surpassing both sentence-level and word-level methods (+16.61$\%$ $\sim$ +0.21$\%$).

\subsubsection{Ablation of Top 1-10 Retrieved Images Across Various Baselines.}
\begin{figure}[t]
    \centering
    \begin{center}
    \includegraphics[width=2.7in]{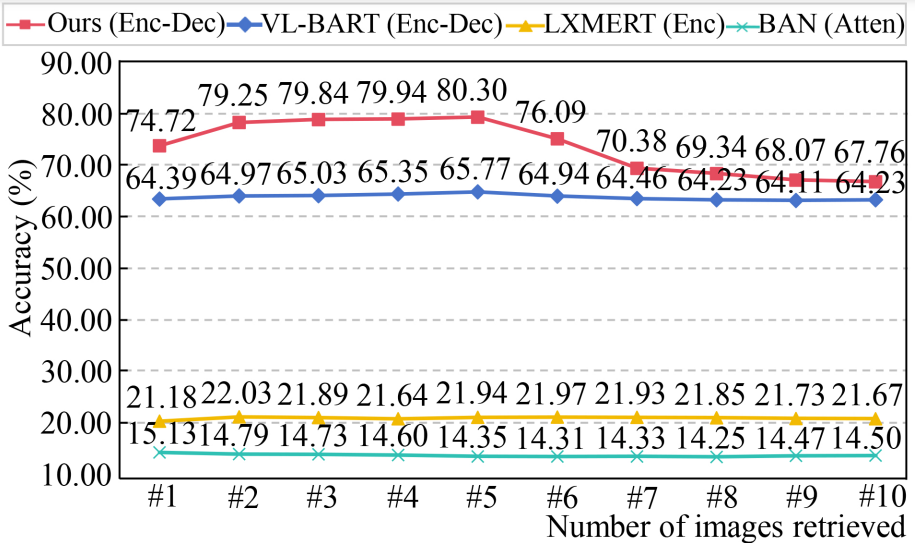}
    \end{center}
    \caption{Performance comparison of our method with various baselines under different numbers of retrieved images.}
    \label{fig:line}
\end{figure}

Figure \ref{fig:line} illustrates the performance across different numbers of retrieved images, 
comparing our approach with various baseline architectures,
including attention-based, encoder-only and encoder-decoder visual-language models, 
all within the ``Introduce Retrieval into QA" category. 
(1) 
Our method consistently outperforms other baselines across the top 1-10 retrieved images in multi-image QA, 
demonstrating its broad applicability. 
Specifically, it exceeds the encoder-decoder baseline VL-BART by +14.53$\%$ $\sim$ +3.53$\%$ across these images, 
underscoring 
the effectiveness of the MHyS-driven paradigm in enhancing both retrieval and QA.
(2) 
Our method's performance peaks at 80.30$\%$ with five images, suggesting that both too few and too many images are not optimal for answering questions.
(3) Simpler models, like attention-based BAN and encoder-only LXMERT, struggle with multi-image QA, indicating the need for more sophisticated models. 
Our encoder-decoder approach outperforms BAN by +65.17$\%$ and LXMERT +58.27$\%$. 
LXMERT performs better with two images, 
while BAN is more effective with one, likely due to their design differences. 
LXMERT excels at modeling multiple object relationships, 
while BAN focuses on bilinear fusion of a single image and text.

\subsubsection{Ablation of Retrieval-based Single-image QA.}
\begin{table}[t]
  \centering
  \fontsize{220pt}{220pt}\selectfont
  \resizebox{1.0\columnwidth}{!}{
    \begin{tabular}{l||c|ccccc}
    \toprule
    Models & All & Att  & Col & Cou & Rel & Sha \\
    \midrule
    \rowcolor{mygray}
    \textit{Retrieve then answer} &      &       &       &       &       &  \\
    mPLUG \citeyearpar{li2022mplug}-BAN \citeyearpar{kim2018bilinear} &  14.19 & 0.0 & 18.45 & 0.31 & 6.74 & 38.13 \\
    mPLUG \citeyearpar{li2022mplug}-VisualBERT \citeyearpar{li2019visualbert}&   17.41     &  0.00 &  28.30 &  0.29 &  10.10  & 40.50 \\
    mPLUG \citeyearpar{li2022mplug}-LXMERT \citeyearpar{tan2019lxmert}&   17.62 &  0.00     &   26.62 &  0.26 &  12.04 & 42.32 \\
    mPLUG \citeyearpar{li2022mplug}-VLBART \citeyearpar{cho2021unifying} &  56.43 &  72.69 &47.57 & 54.37 & 6.07 &73.92\\
    \midrule
    \rowcolor{mygray}
    \textit{Introduce retrieval into QA} &      &       &       &       &       &  \\
    CLIP \citeyearpar{radford2021learning}-BAN \citeyearpar{kim2018bilinear} &   15.13 &  0.00 & 18.3&  0.32& 14.34&  39.31\\    
    CLIP \citeyearpar{radford2021learning}-VisualBERT \citeyearpar{li2019visualbert} &  19.00 &0.00 & 27.31 & 0.39 & 21.89     & 43.48 \\    
    CLIP \citeyearpar{radford2021learning}-LXMERT \citeyearpar{tan2019lxmert} &  21.18 & 0.00 & 33.62& 0.37 & 25.22 &45.03\\
    CLIP \citeyearpar{radford2021learning}-VLBART \citeyearpar{cho2021unifying}  &  64.39 & 78.85 & 63.48 & 53.35 & 14.73 & 84.06 \\
    \hline
    \textbf{Ours} &  \textbf{74.72} & \textbf{79.01} &\textbf{74.23}  &\textbf{67.96}  &\textbf{33.57}  &\textbf{94.56}  \\
    \bottomrule
    \end{tabular}%
    }
  \caption{Comparison with state-of-the-art approaches on the retrieval-based multi-image QA dataset using the top-1 retrieved image.}    
  \label{tab:addlabel4}%
\end{table}%

Table \ref{tab:addlabel4} reports the comparison of our method with 
state-of-the-art approaches using top-1 retrieved image,
categorized into ``retrieve then answer" and ``Introduce  retrieval into QA".
We can observe two key points:
(1) Our method achieves improvements of +18.29$\%$ over the ``retrieve then answer" SoTA mPLUG-VLBART and +10.33$\%$ over the ``Introduce  retrieval into QA" SoTA CLIP-VLBART, 
demonstrating its effectiveness and generality.
(2) Table \ref{tab:addlabel1} indicates that  two-stage ``retrieve then answer" methods generally surpass ``Introduce  retrieval into QA" methods in multi-image QA. 
However, Table \ref{tab:addlabel4} reveals the opposite result for using top-1 retrieved image. 
Concretely, the ``Introduce  retrieval into QA" SoTA CLIP-VLBART outperforms the ``retrieve then answer" SoTA mPLUG-VLBART by +7.96$\%$, 
suggesting that the ``Introduce  retrieval into QA" paradigm is more robust with limited visual information and  is becoming the dominant trend in tackling this task.

\subsubsection{Ablation of MLLMs' Performance.}

\begin{table}[t]
  \centering
  \fontsize{100pt}{100pt}\selectfont
  \resizebox{1.0\columnwidth}{!}{
    \begin{tabular}{l|c|c|ccccc}
    \toprule
    Models &\textit{FT} or \textit{ZS} & All & Att  & Col & Cou & Rel & Sha \\
    \midrule
    \rowcolor{mygray}
    \textit{Multiple-images Retrieval then answer} &  &        &       &       &       &       &  \\
    mPLUG \citeyearpar{li2022mplug}-QWENVL \citeyearpar{bai2024qwenvl}& \textit{ZS} &      6.70 &   0.93    &  13.69     &   8.75    &  2.12     &  4.03\\    
    mPLUG \citeyearpar{li2022mplug}-mPLUG-Owl2 \citeyearpar{ye2024mplug}  & \textit{ZS}&    35.32   &   54.20    &   42.81    &   33.08    &  7.39     & 25.75 \\    
    mPLUG \citeyearpar{li2022mplug}-QWENVL \citeyearpar{bai2024qwenvl}& \textit{FT} & 54.43      &  75.71     &  48.54     &  45.97     &   4.57    & 71.12 \\
    \midrule
    \textbf{Ours} \textit{Introduce  retrieval into QA}&  & \textbf{80.30} & \textbf{78.51} & \textbf{79.45} & \textbf{80.52} & \textbf{40.44} & \textbf{97.85} \\   
    \midrule
    \midrule
    \rowcolor{mygray}
    \textit{Single-image  Retrieval then answer}&  &        &       &       &       &       &  \\    
    mPLUG \citeyearpar{li2022mplug}-QWENVL \citeyearpar{bai2024qwenvl} & \textit{ZS}&     8.57  &  11.68     & 12.92      &  8.04     & 4.06      &4.00  \\  
    mPLUG \citeyearpar{li2022mplug}-LLaVA-1.5 \citeyearpar{liu2024improved}&\textit{FT}  &34.70   &   51.24    &  44.58     &  32.42     &  9.02     &  23.76   \\  
    mPLUG \citeyearpar{li2022mplug}-LLaVA-1.5 \citeyearpar{liu2024improved}&\textit{ZS} &  34.85     &   51.34    &  44.45     & 32.48      &   8.76    & 24.48 \\   
    mPLUG \citeyearpar{li2022mplug}-mPLUG-Owl2 \citeyearpar{ye2024mplug} & \textit{ZS}&   35.09    &  53.28     &    42.78   &  33.28     &  7.08     & 25.45 \\    
    mPLUG \citeyearpar{li2022mplug}-QWENVL \citeyearpar{bai2024qwenvl}&\textit{FT} & 59.73      &   77.73    & 54.89      &  57.43     &  4.91     & 73.72 \\
    \midrule
    \textbf{Ours} \textit{Introduce  retrieval into QA} & & \textbf{74.72} & \textbf{79.01} &\textbf{74.23}  &\textbf{67.96}  &\textbf{33.57}  &\textbf{94.56}  \\      
    \bottomrule
    \end{tabular}%
    }
  \caption{Comparison with the state-of-the-art MLLMs. ``\textit{FT}" represents answers generated after fine-tuning the MLLMs. ``\textit{ZS}" denotes answers generated directly in a zero-shot manner by MLLMs.} 
  \label{tab:addlabel10}%
\end{table}%


Table \ref{tab:addlabel10} represents the comparison of our method with state-of-the-art multimodal large language models (MLLMs). 
These MLLMs, trained on extensive datasets, exhibit strong  generation capabilities
and are utilized in the two-stage ``retrieve-then-answer" pipeline with both fine-tuning and zero-shot ways.
For retrieval, we use mPLUG \cite{li2022mplug},
known for its cross-modal skip-connections and top performance in image-text retrieval. 
We gain:
(1)  Although our method is part of the``Introduce retrieval into QA" category,
it outperforms the fine-tuned MLLM QWENVL \cite{bai2024qwenvl} by +25.87$\%$ in multi-image QA and +14.99$\%$ in single-image QA, 
demonstrating 
the effectiveness of our MHyS-driven paradigm.
(2) Both fine-tuned and zero-shot MLLMs perform better on single-image QA than in multi-image QA, 
which contrasts with typical vision-language pre-trained models.
This might be because mPLUG, despite its SoTA performance in single-image caption-based retrieval, struggles with multiple question-related images.
MLLMs have not seen multi-image QA datasets during pretraining, making their vision-language alignment less effective for this task.
There is an urgent need for a MLLM capable of handling retrieval-based multi-image QA. 

\subsection{Analysis}

\subsubsection{Performance with Different Question Types.}
\begin{figure}[t]
    \centering
    \begin{center}
    \includegraphics[width=3.3in]{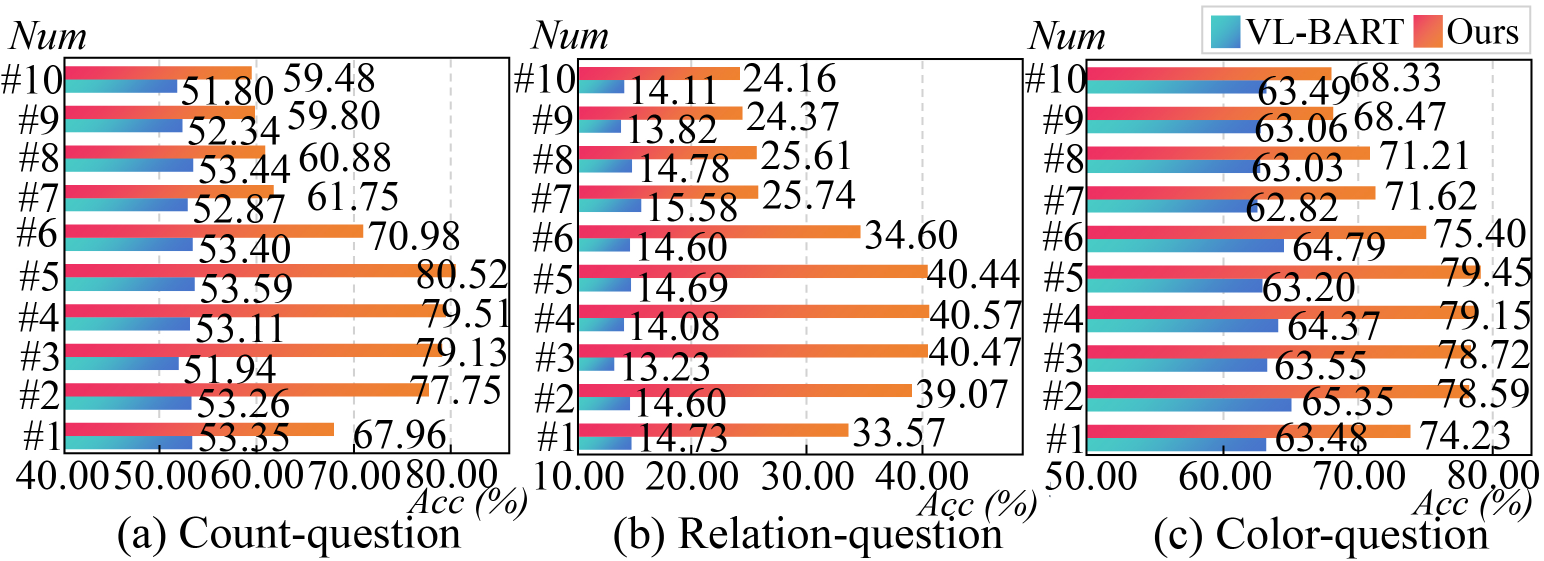}
    \end{center}
    \caption{Performance with different question types.}
    \label{fig:chart}
\end{figure}

Figure \ref{fig:chart} provides the  
quantitative result across various question categories,
from which we can observe:
(1) \textbf{Our method significantly surpasses the baseline in multiple areas}, including counting, relation and color questions, with varying numbers of retrieved images.
Specifically, we find the improvements in count question (+26.93$\%$ $\sim$ +7.68$\%$), relation question (+25.75$\%$ $\sim$ +10.05$\%$) and color question (+16.25$\%$ $\sim$ +4.84$\%$).
(2) \textbf{Our method is more proficient at  solving count question}.
(80.52$\%$ \textit{vs}  53.59$\%$)
This might be because description-form MHyS offers more specific contextual information,
while question-form MHyS focuses on key object, enhancing robustness in solving count question.

\subsubsection{Qualitative Analysis.}

\begin{figure}[t]
    \centering
    \begin{center}
    \includegraphics[width=3.2in]{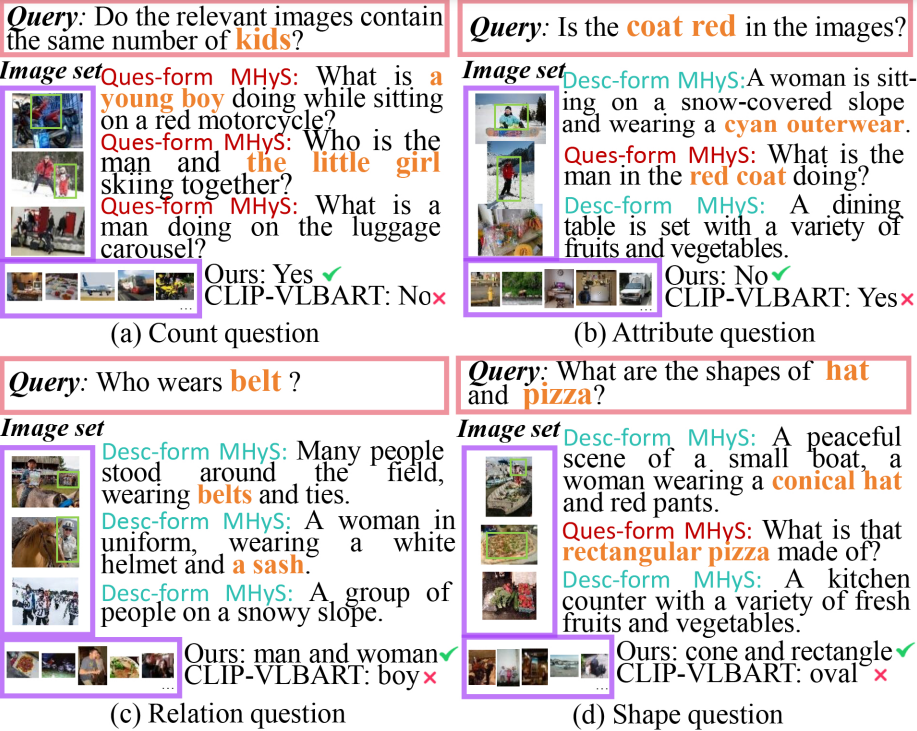}
    \end{center}
    \caption{Qualitative comparison between our method and the baseline.}
    \label{fig:chart}
\end{figure}

To visually demonstrate the effectiveness of our  
MHyS-driven paradigm (``Introduce retrieval into QA"),
we present four examples (count, attribute, relation and shape question) in Figure \ref{fig:chart}.
Figure \ref{fig:chart}(a) illustrates \textbf{question-form MHyS effectively addresses count question}. 
For the query (question) ``Do the relevant images contain the same number of kids?", 
the question-form MHyS includes ``a young boy" and ``the little girl",
aiding in retrieving (query word ``kids”) and comparing their numbers.
Figure \ref{fig:chart}(b) clarifies \textbf{both question-form and description-form MHyS collaboratively enhance the ability to answer attribute question}. 
For the query ``Is the coat red in the images?",
the question-form MHyS contains ``red coat", 
aligning with the query ``coat red", 
while the description-form MHyS in the first image provides a contrasting clue ``cyan outerwear",
helping determine the correct answer ``No".
Figure \ref{fig:chart}(c) demonstrates \textbf{the robustness of MHyS in solving relation question}. 
For the query ``who wears a belt?",
the description-form MHyS identifies individuals like ``people wearing belts" and ``woman wearing a belt", 
despite the presence of noisy objects.
Figure \ref{fig:chart}(d) highlights that \textbf{MHyS provides valuable answer-related clues for solving complex open-ended shape question}.
For the query ``What are the shapes of the hat and pizza?", 
the description-form MHyS identifies ``a conical hat" and the question-form MHyS offers  ``rectangular pizza",
enhancing the model's effectiveness.
In comparison, the baseline CLIP-VLBART without MHyS fails to provide correct answers.

\section{Conclusion}

In this paper, 
we propose using multimodal hypothetical summary (MHyS) to 
introduce and reference retrieval into QA.
This paradigm 
replaces 
queried real images with question-form and description-form MHyS in retrieval, 
enabling transformation into text-to-text retrieval.
Incorporating multiple perspectives, MHyS provides question-attended evidence for QA. 
We employ a coarse-to-fine strategy to filter irrelevant details via sentence- and word-level similarity. 
Besides, contrastive enhancement loss is used to further align MHyS with the query.
Our approach achieves a 3.7$\%$ improvement over state-of-the-art methods on RETVQA and a 14.5$\%$ gain over CLIP.
We hope our work inspires more researchers to explore 
retrieval-based multi-image QA.

\section{Acknowledgments}
This work was supported by the National Natural Science Foundation of China (Nos. 62472419, 62472420, 62072212, 62302218),  the Development Project of Jilin Province of China (No. 20220508125RC).

\bibliography{aaai25}

\end{document}